\documentclass{article}

\PassOptionsToPackage{numbers}{natbib}

\usepackage[final]{neurips_2022}

\usepackage{amsthm}         %
\usepackage[utf8]{inputenc} %
\usepackage[T1]{fontenc}    %
\usepackage[colorlinks,linkcolor=blue,citecolor=blue]{hyperref}       %
\usepackage{url}            %
\usepackage{booktabs}       %
\usepackage{amsfonts}       %
\usepackage{nicefrac}       %
\usepackage{microtype}      %
\usepackage{xcolor}         %
\usepackage{enumitem}       %
\usepackage{framed}         %

\usepackage{array}
\newcolumntype{P}[1]{>{\centering\arraybackslash}p{#1}}
\usepackage{multirow}

\usepackage{caption}
\usepackage{subcaption}
\usepackage{xspace}
\usepackage{wrapfig}
\usepackage{graphicx}
\usepackage{MnSymbol}
\usepackage{ifthen}
\usepackage{cleveref}
\usepackage{authblk}
\newboolean{include-notes}
\setboolean{include-notes}{true}

\usepackage{algorithm}
\usepackage{algpseudocode}

\DeclareRobustCommand{\MicahComment}[1]{\ifthenelse{\boolean{include-notes}}
 {{\color{cyan}M: #1}}{}}
\DeclareRobustCommand{\DavidComment}[1]{\ifthenelse{\boolean{include-notes}}
 {{\color{orange}D: #1}}{}}
 \DeclareRobustCommand{\abnote}[1]{\ifthenelse{\boolean{include-notes}}
 {{\color{blue}AB: #1}}{}}
\DeclareRobustCommand{\adnote}[1]{\ifthenelse{\boolean{include-notes}}
 {{\color{red}A: #1}}{}}
\DeclareRobustCommand{\todo}[1]{\ifthenelse{\boolean{include-notes}}
 {{\color{red} #1}}{}}

\newcommand{\relic}{CLRVis\xspace} %
\newcommand{\tsne}{t-SNE\xspace}
\newcommand{\DRLHP}{DRLHP\xspace}

\newcommand{\prg}[1]{\noindent\textbf{#1}}

\def\addvalue#1#2{\expandafter\gdef\csname my@data@\detokenize{#1}\endcsname{#2}}
\def\usevalue#1{%
  \ifcsname my@data@\detokenize{#1}\endcsname
    \csname my@data@\detokenize{#1}\expandafter\endcsname
  \else
    \expandafter\ERROR
  \fi
}

\title{Time-Efficient Reward Learning \\via Visually Assisted Cluster Ranking}%

\author[]{\textbf{David Zhang}}
\author[]{\textbf{Micah Carroll}}
\author[]{\textbf{Andreea Bobu}}
\author[]{\textbf{Anca Dragan}}
\affil[]{UC Berkeley\thanks{Corresponding authors: \{davidwzhang, mdc, abobu\}@berkeley.edu}}

\begin{document}

\maketitle

\begin{abstract}
One of the most successful paradigms for reward learning uses human feedback in the form of comparisons. Although these methods hold promise, human comparison labeling is expensive and time consuming, constituting a major bottleneck to their broader applicability. Our insight is that we can greatly improve how effectively human time is used in these approaches by batching comparisons together, rather than having the human label each comparison individually. To do so, we leverage data dimensionality-reduction and visualization techniques to provide the human with a interactive GUI displaying the state space, in which the user can label subportions of the state space. Across some simple Mujoco tasks, we show that this high-level approach holds promise and is able to greatly increase the performance of the resulting agents, provided the same amount of human labeling time.
\end{abstract}

\section{Introduction}

Learning reward functions from human feedback has been a successful paradigm in both sequential decision-making problems \cite{abbeel2004apprenticeship,ziebart2008maximum,finn2016gcl,bajcsy2017phri,christiano2017preferences,brown2019extrapolating}, enabling AI agents to learn behaviors that are challenging, or perhaps even impossible, to specify manually. Although these methods hold promise, the main bottleneck to applying them broadly in real-world scenarios is the amount of human effort they require, sometimes requiring hours of uninterrupted human attention and interaction~\cite{christiano2017preferences}. %

To make it easier for the human to provide input, many recent approaches use a flexible and human-friendly type of reward feedback in the form of comparisons between trajectory snippets~\cite{christiano2017preferences}.
However, this approach is still too slow for practical use because human time is not utilized very effectively: every time the AI agent interacts with the human, it typically receives feedback for a single query at a time. %

\begin{figure*}[h!]

  \vspace{-0.6em}
  \centering
  \includegraphics[width=\textwidth]{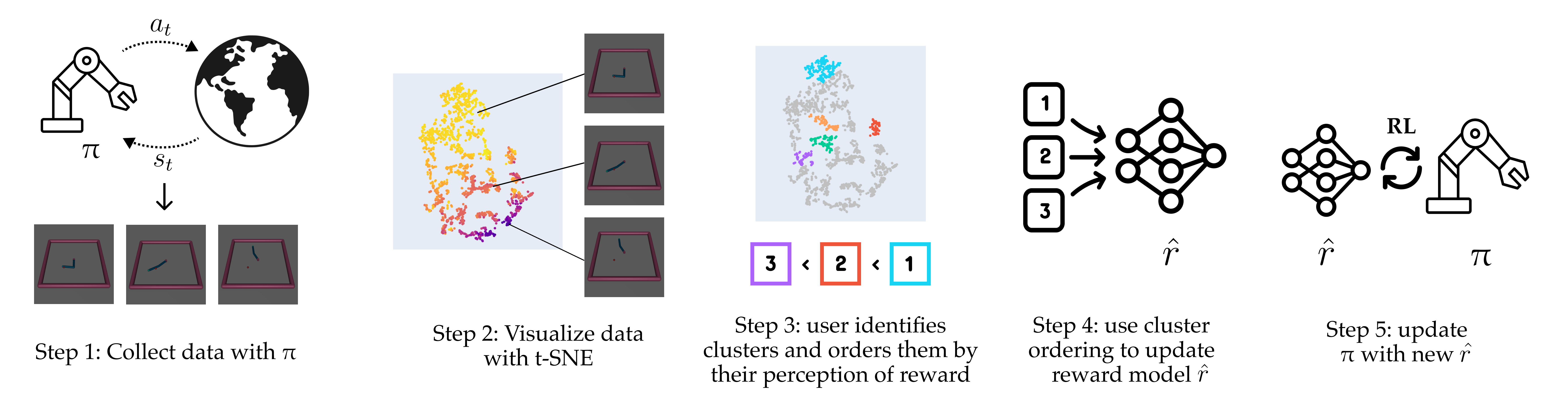}
  \caption{\textbf{The \relic algorithm.} Our method consists of several stages: we collect environment interactions using our current policy $\pi$ and show them to the user through a \tsne visualization. The user then provides feedback in the form of cluster orderings, which are then used to update the reward model $\hat{r}$; in turn, the reward model $\hat{r}$ is then used to update the policy $\pi$ and the process repeats.
  }
  \label{fig:method}
\end{figure*}

Recent techniques based on active learning attempt to further alleviate the burden on the human by shifting it onto the robot: rather than randomly select which queries to ask the human for feedback on, the robot itself tries to come up with the most informative ones, in order to be efficient with the human's time.
In this project, we are interested in a complementary question: rather than design metrics for having the robot decide which queries are most informative, we want to build tools to assist humans in providing feedback on \textit{multiple} related queries at once.

For instance, imagine you were given an interactive visualization of the state space for a self-driving task, in which similar states are placed close together: in the top-right, you hover over various states and by looking at the states, you can tell they all are bendy roads; you zoom in and see subclusters of states in which the car is swerving in the other lane, or well-centered in its lane. This could enable the human to provide feedback on many semantically similar states at once, at the level of granularity that the human deems to be most appropriate. 
Using a visual interface of this kind, it seems like the human would be able to give quick, rich feedback on entire sets of similar states at once.

While this kind of tool might seem taken out of a sci-fi hologram scene, in this work we set out to build a similar type of system. We test such an approach can, in fact, make more efficient use of the human's time than current state of the art methods. 
Our insight is that enabling the human to provide input over large swaths of the state space at once relies on how that state space is visually presented to them: if states that behave similarly are visualized close together, then the person should be able to more easily think of them similarly and group them together.
As such, we use high-dimensional data visualization techniques~\cite{vandermaaten2008tsne} to cluster states based on both their visual and reward characteristics.
We make use of recent contrastive techniques~\cite{oord2018CPC,laskin2020curl,chen2020simclr} to learn an embedding that maps visually similar states near each other, and combine it with  the current trained reward embedding to visualize states close both in reward and appearance near each other.

We show in our experiments that our method (which we call \relic) requires up to 7x less human time than a \DRLHP-like baseline across some Mujoco reward-learning tasks. We validate this with some real human data and a suite of simulated human-feedback experiments.

\section{Related Work}

\prg{Reward Learning from Human Input.}
A popular paradigm for learning robot behaviors from human input is inverse reinforcement learning (IRL), where the robot seeks to extract a reward function capturing why specific behaviors may be desirable. Traditionally, IRL relies on human demonstrations to recover the reward function responsible for the human's input~\cite{abbeel2004apprenticeship,ziebart2008maximum,argall2009survey, fu2018learning,finn2016gcl,wulfmeier2016maxentirl}.
Recent research goes beyond demonstrations, utilizing other types of human input to learn rewards, such as corrections~\citep{bajcsy2017phri}, comparisons~\citep{wirth2017survey,christiano2017preferences}, labels~\cite{Warnell2018DeepTI,reddy2019learning}, rankings~\citep{brown2019extrapolating}, or some combination~\cite{Ibarz2018RewardLF,Palan2019LearningRF,mehta2022unified}. Comparisons have been especially successful in recent NLP approaches \cite{Ouyang2022TrainingLM,Stiennon2020LearningTS, Bai2022TrainingAH}.

The issue that persists across all these methods is that the human still bears most of the labeling burden. Active learning (AL) attempts to alleviate some of this burden on the human by shifting it onto the robot: to reduce the total amount of necessary human labels, the robot tries to come up with the most informative queries to ask the human for help with. Robot learning methods have investigated a variety of metrics the robot can optimize to select the best queries, from maximizing uncertainty estimates~\cite{singh2019activeRL}, minimizing performance risk~\cite{brown2018risk}, maximizing exploration~\cite{biyik2019batch}, minimizing human answer uncertainty~\cite{biyik2020easyQ}, or a combination of the above~\cite{reddy2019learning,bobu2022perceptual}.
While researchers in data visualization and labeling find that active learning is an effective way to correctly label data, it is not necessarily the most \textit{efficient}~\cite{Bernard2018ComparingVL}. This is in part due to the cold start problem of AL, where without a good initial model all of these selection strategies start poorly. \cite{Bernard2018ComparingVL} finds that user-interactive labeling can be much more efficient than designing active learning metrics for collecting better reward queries from the human -- following this works' footsteps, we are interested constructing tools that assist humans in labeling more reward data faster.

\prg{AI Assistance in User-Interactive Feedback.}
Prior work in user-interactive labeling has used semi-supervised~\cite{desmond2021aiassisted} or self-supervised~\cite{wang2022pico} models to predict the most likely label associated with the query. This strategy can assist the user in labeling individual queries faster, but the user still has to label each data point one at a time. Instead, more recent approaches propose using visual interfaces to help humans label more data, which can even enable unifying visual assistance with active learning~\cite{bernard2018VIAL}. Based on this, \cite{grimmeisen2020pointing} proposes a visual interactive labeling approach where the human can select clusters via a lasso tool, after which a model recommends a label for each cluster. Then, the interface allows the human to deselect individual points that don't match the cluster's overall label. Instead, \cite{blei2020clusterclean} use a clustering algorithm to propose clusters automatically to the human, who can manually remove instances that don't belong to each cluster. Both methods have primarily been demonstrated on MNIST and only work on categorical variables, whereas the reward learning scenarios we are interested in require predicting reward as a continuous variable. Since humans are notoriously noisy when labeling continuous variables~\cite{Braziunas2008elicitation}, we similarly look at visual interfaces that use clustering to assist the user, but opt for ranking clusters instead of labeling them.

\prg{High-dimensional Data Visualization.} As we just saw, one way to construct better tools for labeling data is to construct better ways to visualize that data.
\tsne is a non-linear dimensionality reduction technique commonly used for visualizing high dimensional data in an easy and intuitive way~\cite{vandermaaten2008tsne}. Most relevant to our work is its widespread use in reinforcement learning as a tool for visually evaluating the quality of the learned RL agent embeddings~\cite{Mnih2015HumanlevelCT,zahavy2016graying,aytar2018playing,annasamy2019interpretability}. In these works, the intuition behind using \tsne is that if an agent learned a good embedding for Q-value, reward, etc., states that are embedded similarly -- i.e. have similar Q-values or rewards -- will appear close in \tsne space, even if they are perceptually dissimilar. We use a similar visualization technique, but we differ in that we don't only use it for qualitative evaluation purposes at the end of training an RL agent. Rather, we seek to recompute the \tsne visualization at every labeling iteration, after having both experienced more states and retrained the embedding.

The above methods typically apply \tsne on states sampled during the fully trained RL agent's execution of its policy~\cite{Mnih2015HumanlevelCT} or during the agent's entire training experience~\cite{zahavy2016graying}. Since we recompute \tsne at every iteration, we instead only have access to states from the current iteration policy or from prior policies. In the future, we could assume a small amount of successful task demonstrations and sample states according to a curriculum~\cite{florensa2017curriculum} or along an expert demonstration of the task~\cite{Salimans2018LearningMR}.
Furthermore, these methods have either visualized the raw pixel representation or the final layer Q-value network embedding of the sampled states. Since we are primarily interested in relieving the visual labeling burden from the human, we also train a contrastive network~\cite{oord2018CPC,chen2020simclr,stooke2021decoupling,laskin2020curl}, which seeks to learn visually discriminative embeddings. We concatenate the learned reward embedding and the contrastive one, and compute the \tsne visualization from the concatenated vector.
Prior work has looked at constructing representation embeddings that result in visualizations more interpretable to the human~\cite{hilgard2021MoM}, but there the assumption is that the robot is the teacher and the human is the student trying to learn most effectively from that visualization.

\section{\relic}

We call our methodology \textbf{CL}usters \textbf{R}ankings with \textbf{Vis}ual assistance (\relic). %
On a high-level, our method consists of iteratively showing a person visualizations of subsets of the state space, which can be used as an interface to efficiently provide additional reward labels (see \Cref{fig:method}). %
Every iteration of \relic can be broken down into various steps:
\begin{enumerate}
    \item \textbf{State space sampling:} we sample some states using our current policy $\pi$.
    \item \textbf{Visualizing states and obtaining cluster rankings:} we visualize states sampled in step 1 with \tsne, and have the user identify $M$ clusters of similar states through a graphical interface, and provide a ranking of such $M$ clusters in terms of increasing reward.
    \item \textbf{Converting cluster rankings to pairwise state comparisons:} for each pair of clusters in our cluster ranking, we can then consider all individual pairs of states and treat them as single-state comparisons (if cluster $A$ was ranked higher than cluster $B$, we can treat this as equivalent to stating that state $a\in A$ has higher reward than state $b \in B$). This gives us on the order of $O(M^2 S^2)$ comparisons (if the size of the clusters was fixed to be $S$), where the $M^2$ term comes from transforming cluster rankings into cluster comparisons, and the $S^2$ term comes from comparing two clusters.
    \item \textbf{Updating reward model:} with these $O(M^2 S^2)$ comparisons, we can update our running reward model $\hat{r}$. 
    \item \textbf{Updating policy:} with our now-improved reward model $\hat{r}$, we can train $\pi$ further, and return to step 1. For a schematic description of the algorithm, see \Cref{alg:method}.
\end{enumerate}

In short, providing cluster reward-rankings with the aid of an appropriate visual interface should not take many times longer than comparing individual states. However, such cluster rankings could contain up to $O(M^2 S^2)$ more information than an individual cluster comparison. While we might not expect a $O(M^2 S^2)$ speedup in human-time in practice (because many of the $O(M^2 S^2)$ individual comparisons might be similar and not very informative), conceptually this approach still seems significantly better than individual state comparisons, which is the current standard approach.

However, the ease (and thus speed) of cluster ranking will crucially depend on how interpretable the state space visualization is: are highly dissimilar states next to each other? Do states in similar regions of the visualization have similar reward (and thus labeling them as a cluster is easy)? For the purpose of encouraging visualizations which are conducive to ease of cluster labeling, we combine two approaches: 1) we train a contrastive learning embedding for each new set of state images, and 2) after the first update of the reward model $\hat{r}$, we append the reward model embedding to the contrastive learning one to encourage the states that we already suspect have similar rewards to be close together in the state space visualization.

\prg{Reward model update.} To update the reward model $\hat{r}$, we use the dataset $D$ of pairwise comparisons gathered from cluster rankings. Each comparison can be represented as a tuple $(s_0, s_1, y) \in D$, where $s_0$, $s_1$ are the two states, and $y \in \{0, 1\}$ is a label indicating which state has higher reward.
Under the Bradley-Terry model, we can assume our reward model is a score function that predicts the likelihood one state is better than another \cite{10.2307/2334029}.

\begin{equation}
P(s_0 > s_1) = \frac{e^{\hat{r}(s_0)}}{e^{\hat{r}(s_0)} + e^{\hat{r}(s_1)}}
\end{equation}

 We optimize $\hat{r}$ with a cross-entropy loss between the predicted labels and true labels as follows:

\begin{equation}
\text{loss}(\hat{r}) = - \sum_{(s_0, s_1, y) \in D}^{} y \log (P(s_0 > s_1)) + (1-y) \log (1 - P(s_0 > s_1))
\end{equation}

\begin{algorithm}
\caption{The \relic algorithm}\label{alg:cap}
\begin{algorithmic}
\Require an initial policy $\pi^0$, an initial reward model $\hat{r}^0$.
\For{$i$ in $0, \dots, K$}
\State Sample $s_0, \dots, s_N$ from rollouts of $\pi^i$.
\State Visualize $s_0, \dots, s_N$ with \tsne, based on embeddings from $\hat{r}^i$.
\State Human selects clusters of states with similar rewards, and orders them by increasing reward.
\State Obtain $\hat{r}^{i+1}$ by updating $\hat{r}^{i}$ with cluster ordering information.
\State Obtain $\pi^{i+1}$ by training $\pi^i$ with the updated reward model $\hat{r}^{i+1}$.
\EndFor
\end{algorithmic}
\label{alg:method}
\end{algorithm}

\section{Experiments}

\subsection{Experimental Setup}

\prg{Environments.} We test our method on 3 MuJoCo environments -- Swimmer, Reacher, HalfCheetah -- but we modify the goal-task. We do this because, similarly to \cite{christiano2017preferences}, we want to showcase that our method can also work for tasks that we don't have an explicit reward for. However, for the purposes of evaluating \relic, we found it useful to hardcode ground truth reward functions for our new task-goals, even though they were hidden from \relic's learning process. The advantage is that we can use such ground truth reward functions both as a oracle performance indicator, and as a way to ground simulated models of user feedback, which we use in our experiments.

\prg{Modified environment tasks.} For Reacher, we set the goal to get the arm as close as possible to a fixed dot position in the bottom left. For HalfCheetah, we change the goal to have the cheetah stand upright on its head, rather than running forward. For Swimmer, we set the goal to curl up in a horseshoe shape. See \Cref{appendix:environment-details} for more information and for the custom reward functions.

\prg{Mixed policy.} We noticed that \relic would sometimes get stuck in local optima during training (i.e. low-reward policies). We suspected this might be due to low diversity of states in the \tsne visualizations (potentially caused by low entropy of the latest policy, used to collect states for the \tsne). To mitigate this issue, we 
combine a random policy with our current policy while collecting episodes. We start each episode by using our current policy, and switch to a random policy after $T$ timesteps, where $T$ is uniformly drawn from the max possible length of the episode.

\begin{figure*}[t]
  \centering
  \includegraphics[width=0.7\textwidth]{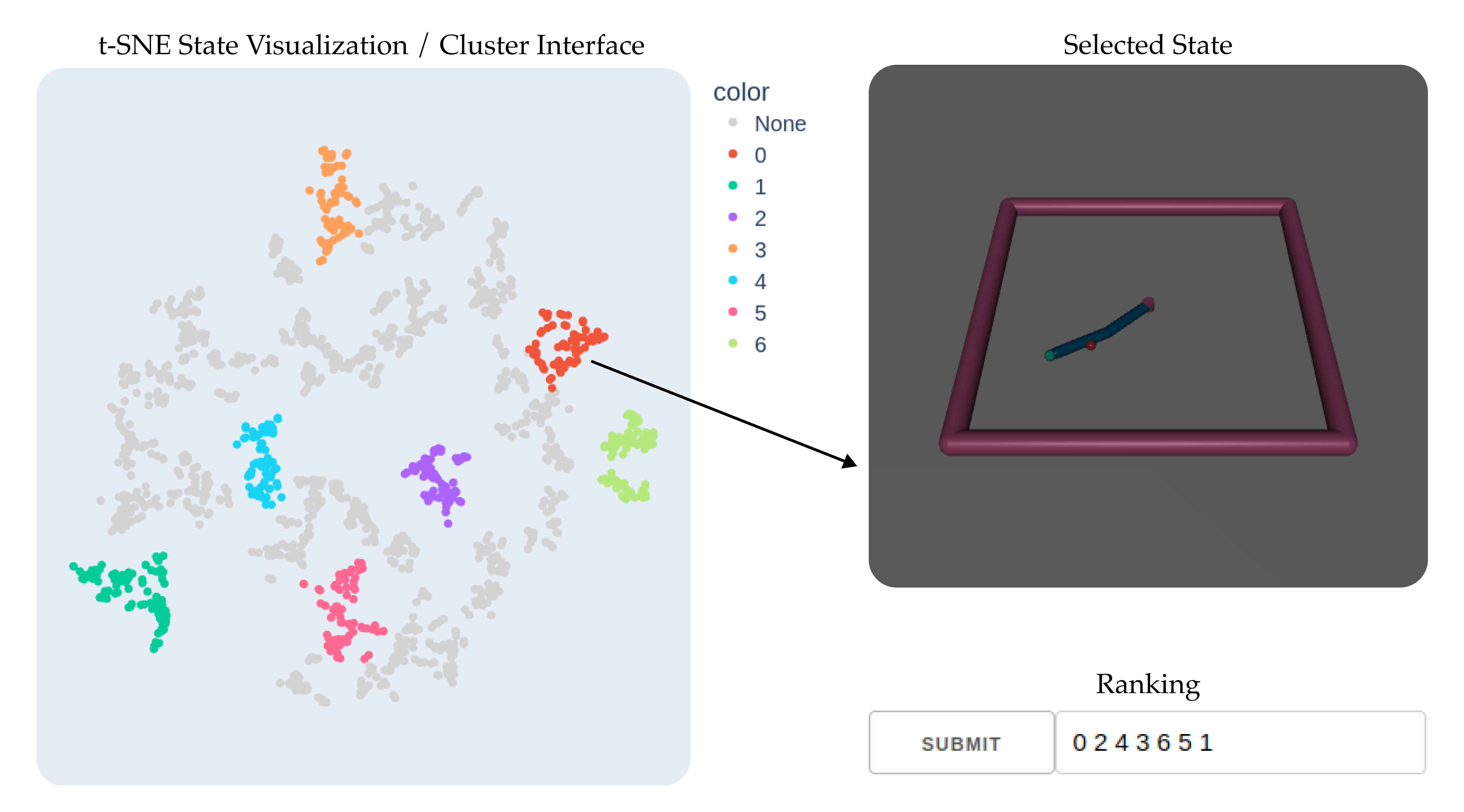}
  \caption{\textbf{State-space Visualization Interface.} A real human can provide reward feedback by selecting clusters in our \tsne state visualization and inputting a ranking. Sample clusters are denoted with different colors. To identify clusters of states with similar rewards, the user hovers over different points on the \tsne which will display an image of the corresponding states on the right.}
  \label{fig:interface}
\end{figure*}

\prg{Visualization and ranking interface.} To collect feedback on reward preferences from the user, we implemented the web interface in \Cref{fig:interface}. At each iteration we display a \tsne visualization of $N$ states, sampled with the mixed policy described above. The user can then hover over each point in the graph, which will display the corresponding image of the environment state. To provide feedback, the user uses a lasso tool to select clusters which they believe to have similar reward values. After selecting $M$ such clusters, the user ranks them by entering their order into a text box. Based on this ranking, we extract pair-wise preferences between states to update our reward model.

\prg{Human-time.} 
In our context of learning a reward function (and optimal policy) from human feedback, the main practical bottleneck is that of human time: how much time are we requesting from humans to provide labels to update our reward model $\hat{r}$? Human time is expensive, and in comparison, the computational cost of updating the policy $\pi$ is negligible. Because of this, we report algorithm performance across different points of human-time, rather than different numbers of iterations.

\prg{Simulated Human -- \relic.} %
Since collecting real human feedback for every experiment we ran was prohibitive, to test our method we implemented a simulated human, which we calibrated to be approximately as good as a real human or worse. We show some comparisons between our simulated human and a real human in \Cref{fig:short-term}. To simulate human cluster choice, we use agglomerative clustering~\cite{scikit-learn} over the \tsne plot, and filter out the clusters that are too small or have high variance relative to other clusters. We then choose $M$ clusters that are equally spaced in terms of average reward. To do this, we first compute $M$ "target values" that are equally spaced from minimum to maximum cluster reward. Then, we select the cluster with average reward closest to each target value. Using the ground truth reward function, we finally rank the clusters from $1$ to $M$ based on their average reward. Note that this simulated human is noisy: the selected clusters will have some variance, and so the cluster rankings will lead to some mis-labeled data, similarly to what would happen with a real human. %
We calibrate how long we expect the simulated human to take per-cluster-ranking for each environment by timing multiple real human \relic runs for a small number of iterations, and take the average of the cluster-ranking times.

\prg{Simulated Human -- \DRLHP.} We also need to simulate human feedback for \DRLHP. To be certain to benefit the baseline method, we assume that human feedback is without noise: every comparison is perfectly labeled, according to the oracle reward function. For each single-state comparison in \DRLHP, we assumed that the human labeling time would be 3 seconds (lower bound from ~\cite{christiano2017preferences}).

\prg{Reinforcement Learning.} To update our policy after learning a reward function, we use PPO for all experiments. While we use images to learn a contrastive representation, both the reward function and policy function use the vectorized observation space as input. For more details, see \Cref{appendix:experimental-details}.

\subsection{Hypotheses}

\prg{Manipulated Variables.}
We test the effect of two different strategies of human feedback collection: 
a version of \DRLHP~\cite{christiano2017preferences} which elicits comparisons between individual states (i.e. in which the trajectory snippet length is equal to 1); and \relic (our method), which instead elicits cluster rankings made through our graphical user interface.%

\prg{Dependent Measures.} To compare each method's performance, we report the oracle \textit{Episode Reward} that each policy type collects after a certain amount of human labeling time.

\prg{Hypotheses.} We have two hypotheses: \textbf{H1.} In low human-time regimes, \relic will recover better rewards than \DRLHP with the same amount of labeling time; \textbf{H2.} In high human-time regimes, \DRLHP will eventually recover similar or better rewards than \relic (because \DRLHP has less mislabeled datapoints).

\subsection{Quantitative Results} 

\begin{figure*}[t]
  \centering
  \includegraphics[width=0.66\textwidth]{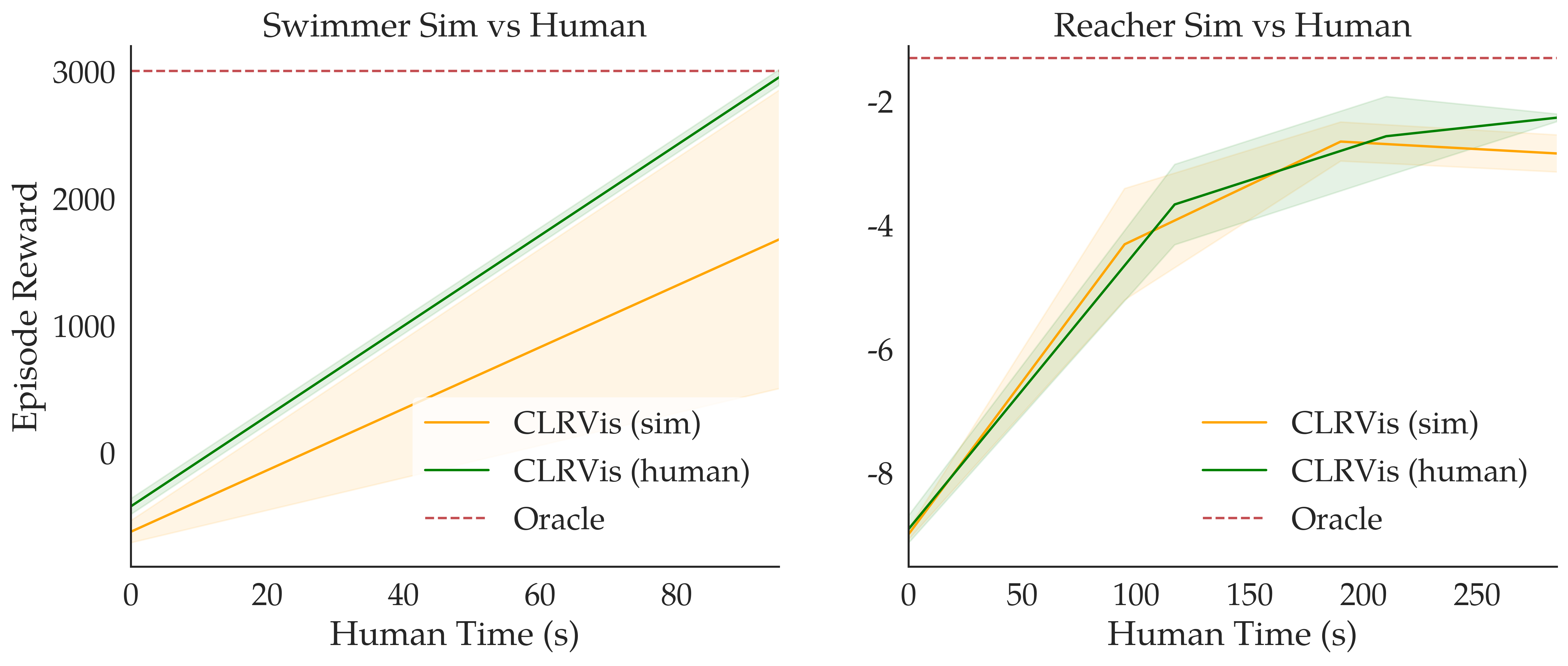}
  \caption{\textbf{\relic simulated vs human.} We run \relic with a real human for Swimmer and Reacher. In Reacher, the simulated human performs roughly the same as the real one. Swimmer only requires one iteration of \relic feedback to converge -- only one seed does poorly in simulation, but the rest match the real human.}
  \label{fig:human-vs-sim}
\end{figure*}

All experiments shown are run with 5 seeds, except for those with real human data, which are run with 3 seeds. All error bars reported are standard errors of the mean across seeds.

\prg{Human model realism.}
For Reacher and Swimmer, we ran \relic with both a real human and our simulated human to teach an agent the intended tasks. We show the results in \Cref{fig:human-vs-sim}, from which we see that the simulated human performance is similar or an underestimate to the real human one. We didn't do this same procedure for Cheetah as because the human time required was too restrictive at this stage of the research. As a quick sanity check, we notice that all models perform similarly when human labeling hasn't yet started (where the x-axis is $0$), so no reward model has yet been learned. While the real human seems to perform significantly better than the simulated human in Swimmer, this is only caused by a single bad seed. The others match the simulated human well.

\begin{figure}[h]
\vskip -0.7em
\centering
\includegraphics[width=\linewidth]{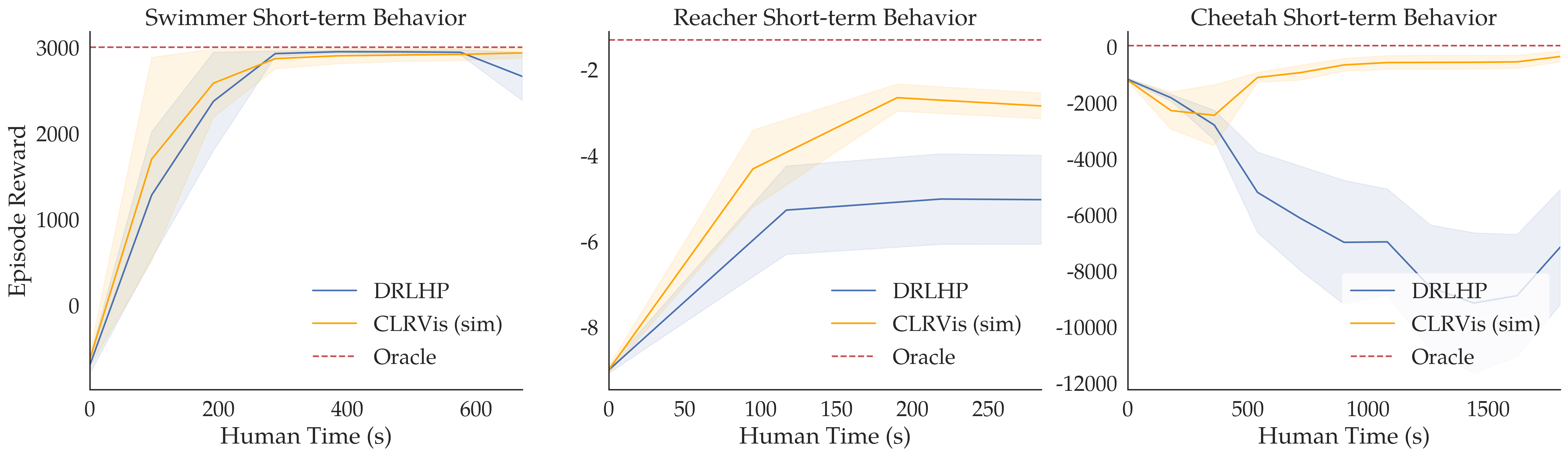}
\caption{\textbf{\relic's time-limited performance.} We see that when human time is constrained, \relic tends to perform significantly better than \DRLHP for the same amount of time, and is able to quickly approach oracle performance across all 3 environments.}
\label{fig:short-term}
\vskip -0.7em
\end{figure}

\prg{\relic's time-efficiency.} 
From \Cref{fig:short-term}, we see that \relic is more time-efficient in the short run and achieves greater reward in the same amount of human time. Despite having some degree of noise in its cluster rankings, \relic is able to get close to the oracle policy with little human effort, performing much better than \DRLHP in Reacher and Cheetah. \relic matches \DRLHP in Swimmer, but both methods converge to the oracle policy. Therefore, these results support \textbf{H1}.

\begin{figure}[h]
\vskip -0.7em
\centering
\includegraphics[width=0.66\linewidth]{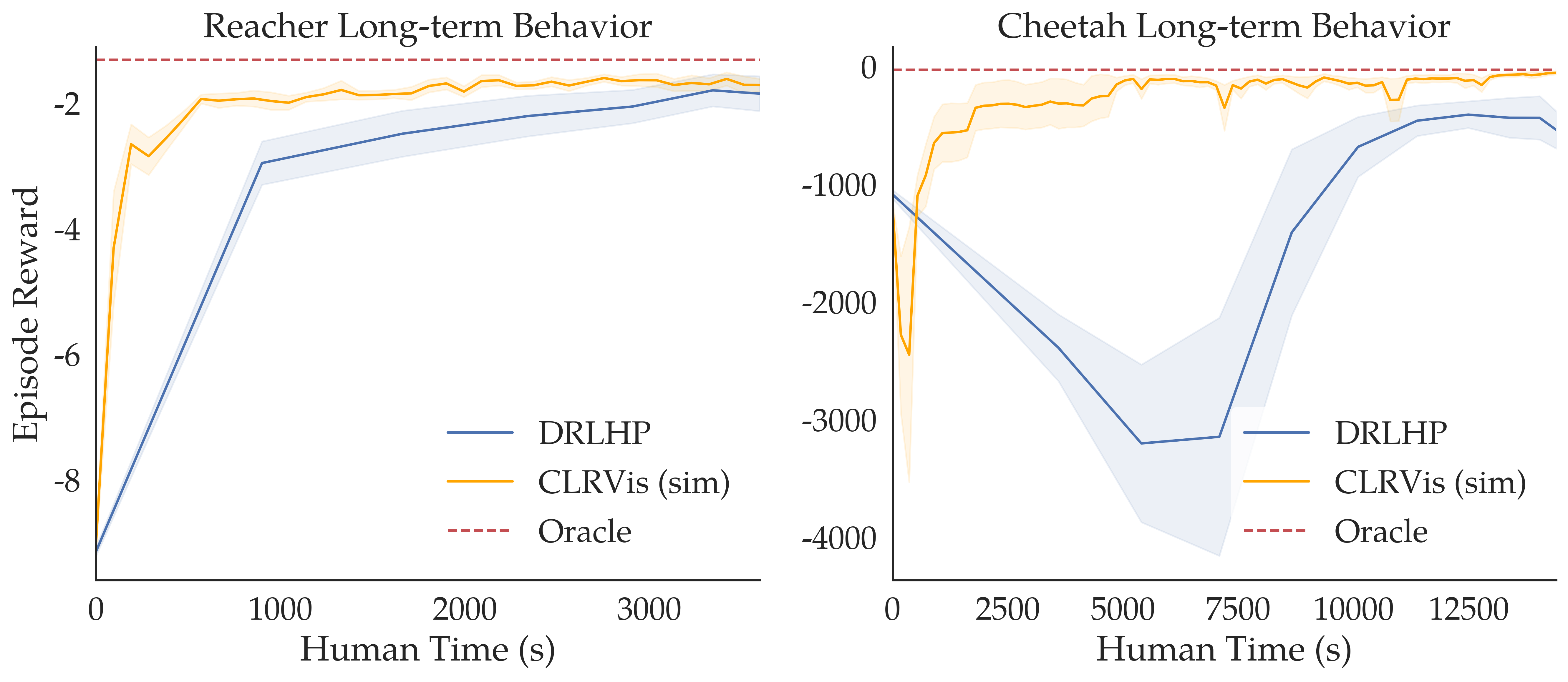}
\caption{\textbf{Long-term results of \relic and \DRLHP.} When we allow \DRLHP to use much more human time to collect comparisons, it's performance approaches that of \relic. However, we still see a significant speedup in human-time when switching from \DRLHP to \relic, roughly 2.5x and 7x for Reacher and Cheetah respectively. }
\label{fig:long-term}
\vskip -1em
\end{figure}

\prg{\DRLHP under unlimited human-time.} From \Cref{fig:long-term}, we see that if one could collect vast amounts of human data, \DRLHP's performance will eventually approach  \relic. However, despite the inherent noise in providing cluster rankings, \DRLHP does not surpass \relic, and in fact does slightly worse, partially supporting H2. Instead, we see a large speedup in human time when using \relic to reach the optimal \DRLHP point. In Reacher, \DRLHP takes 3345 seconds to reach its optimum, while \relic takes 1330 seconds to reach the same point. In Cheetah, \DRLHP takes 12489 seconds to reach its optimum, while \relic only takes 1800. Therefore, we observe roughly a 2.5x and 7x speedup respectively. We do not run long-term experiments on Swimmer because it is able to converge to the oracle policy in the short-term.

For more details about our treatment of human time across our methods and results, see \Cref{appendix:human-timings}.

\subsection{Qualitative Results}

\begin{figure*}[t]
  \centering
  \vskip -1em
  \includegraphics[width=\textwidth]{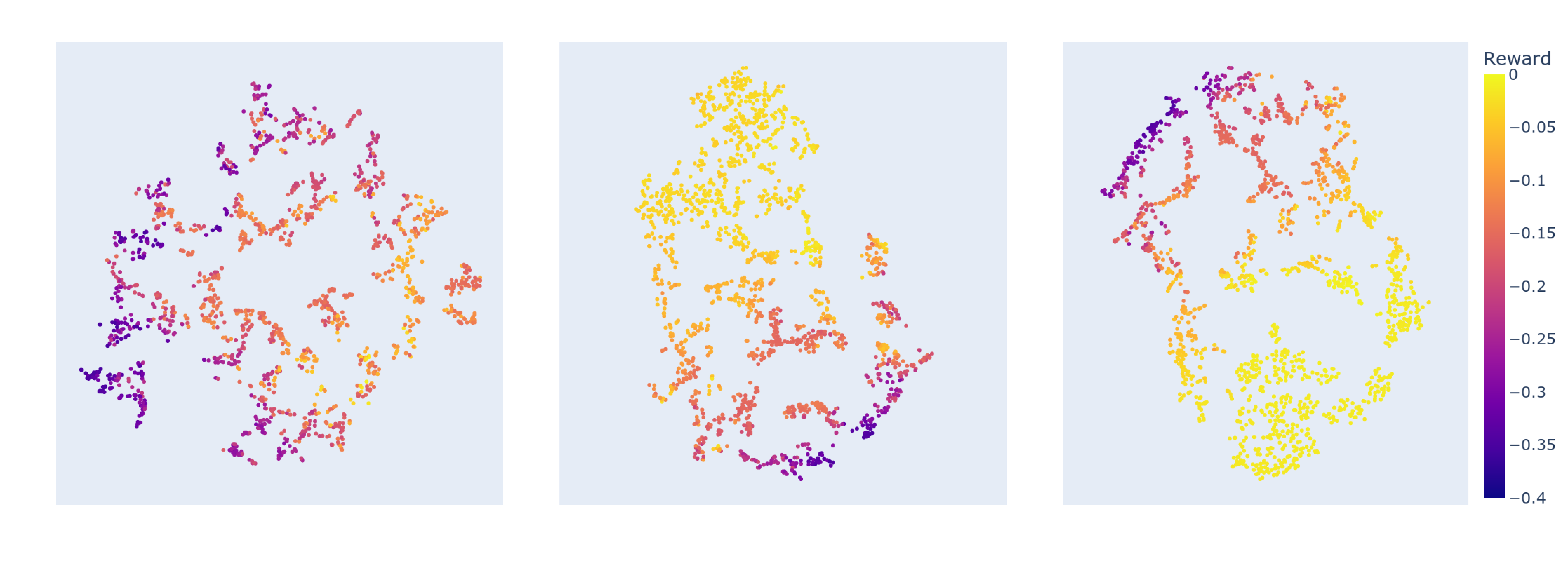}
  \vskip -1em
  \caption{\textbf{} \textbf{\tsne progression.} Over time, the \tsne state visualization becomes easier for a human to cluster. States with similar reward are grouped more closely together as our reward embedding improves.
  }
  \vskip -0.5em
\label{fig:tsne-progress}
\end{figure*}

\begin{figure*}[t]
  \centering
  \includegraphics[width=0.7\textwidth]{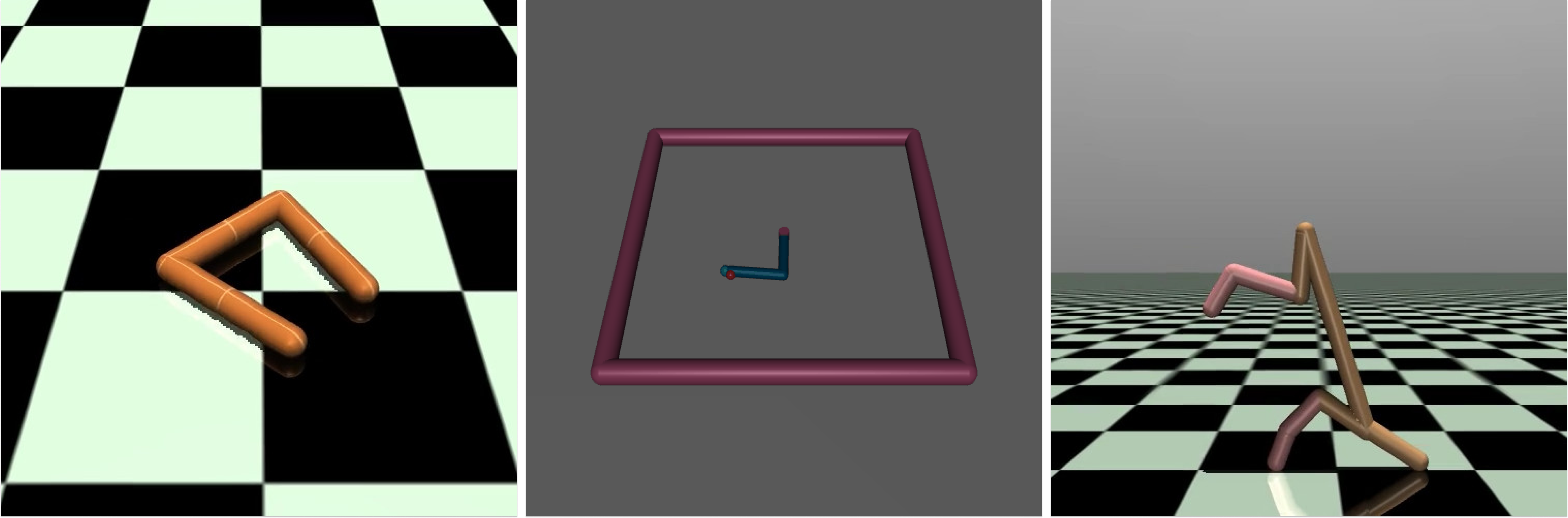}
  \caption{\textbf{} \textbf{\relic end behavior.} \relic is able to learn novel behaviors in the Reacher, Swimmer, and Cheetah environments.
  }
  \label{fig:end-behavior}
\end{figure*}

\prg{\tsne Progression.} As our agent and reward model embedding improves from human feedback, we expect states of similar reward to be more closely grouped in our \tsne visualization. Also, we expect that the average reward in our sampled states will increase due to a better policy.

In \Cref{fig:tsne-progress}, we plot the \tsne visualization of Reacher states after 0, 4, and 8 iterations of \relic feedback. We color the \tsne plots according to the oracle reward values (which would not accessible to the real human labeler). Note how there is a large increase in average reward between iterations 0 and 4, and small increase between iterations 0 and 8. This correlates with \Cref{fig:long-term}, where we see Reacher's episode reward grow sharply towards the beginning of training, and rise slowly later on. In addition, we notice how it is much easier for a human to label cluster rankings in later iterations of feedback. In particular, states of similar reward tend to be become grouped more closely together, and movements in \tsne space correspond to gradual movements along a gradient from high reward to low reward.

\prg{Behavior Examples.} An example of agents reaching their goals in \relic can be found in \Cref{fig:end-behavior}. Videos of agents trained from \relic can be found at \url{https://bit.ly/clrvis-videos}.

\section{Conclusion}\label{sec:conclusion}

\prg{Discussion.} Before deciding to focus on cluster rankings, we also tried other forms of feedback, such as labeling clusters with average reward values or providing individual cluster comparisons. However, we found that labeling clusters with continuous reward values was difficult for a human to do, while cluster rankings performed better than comparisons. Although there was some noise from using clusters instead of individual states, it didn't have a significant effect on our overall results. In our approach, users are also able to actively select the states they want to provide feedback on. As a result, users have more flexibility to choose states in which the expect the reward function to be more nuanced.

This is in contrast to \DRLHP, where users are forced to compare video snippets chosen by the algorithm.

\prg{Limitations.} In this work, we have the users provide feedback over single states (images) instead of trajectory snippets (videos). While comparing images might be easier than comparing video snippets for a human, this limits us to tasks where reward only depends on the current state (a user only needs a single frame to tell if a Cheetah is standing upright). 
Another clear limitation of our work is that most of our experiments are based on simulated humans. %

\prg{Future Work.} In addition to running more extensive real user studies for both \relic and \DRLHP, future work could extend our interface to handle trajectory snippets, enabling to learn more complex tasks that depend on a sequence of images, such as getting the Cheetah to perform a backflip.
In addition, future work might utilize task discovery to pretrain policies in the beginning, rather than rely on a random policy. After doing so, one could fine-tune the model rather than learn from scratch. This is similar to existing NLP methods, where language models are trained on large corpuses of text, and fine-tuned for specific purposes.

\prg{Summary.} Having users provide individual state comparisons for reward feedback can be expensive and take a lot of time. We alleviate this issue by providing an interface where users can provide feedback on large clusters of the state space, which is visualized using \tsne. Generally, we want states with similar reward to be closer to each other, so we make use of contrastive learning and the reward model representation. To prevent our policy from being stuck in local optima, we also introduce the notion of a mixed-policy, where the states shown in our \tsne come from a combination of our current policy and random policy. Through experiments in Mujoco environments, we find that \relic can substantially outperform \DRLHP given the same amount of human feedback time. 

\begin{ack}

We thank the members of the InterACT lab for their invaluable feedback. This work was supported by ONR YIP and the NSF Fellowship.

\end{ack}

\bibliography{refs}
\bibliographystyle{plainnat}

\newpage
\appendix

\section{Environment Details}\label{appendix:environment-details}

\prg{Custom rewards.} Each environment is modified with its own reward function. In general, we remove the torque-based penalties from MuJoCo, which aren't apparent to a human. We also use fixed-length episodes to ensure that termination conditions don't influence our reward function.

Reacher: We use the same reward function as the original environment, but only keep the $reward\_dist$ component and discard $reward\_control$. 

Cheetah: We change the reward function to be $-abs(obs[1] - 1.25)$, where $obs[1]$ is the angle of the main body. The $1.25$ is the target angle for our cheetah, which means we want it to be upright with a slight tilt.

Swimmer: We change the reward function to be $obs[1]*obs[2]$, where $obs[1]$ and $obs[2]$ are the angles of the left and right rotor. We multiply them together to ensure that they are in the same direction, and not in a Z shape instead.

\prg{State and action spaces.} We use the same observation and action spaces as the original Mujoco environment. In general, the observation space contains joint angles, angular and linear velocities, and position coordinates. All of the action spaces are continuous.

\section{Experimental Details}\label{appendix:experimental-details}

\begin{table}
  \caption{\DRLHP Hyperparameters\vspace{8pt}}
  \label{drlhp-params}
  \centering
  \begin{tabular}{ll}
    \toprule
    Parameter & DRLHP \\
    \midrule
    Initial reward training epochs & 200 \\
    Reward training epochs & 2 \\
    Reward batch size & 50 \\
    Reward learning rate & $3*10^{-4}$ \\
    Label schedule & Hyperbolic \\
    \bottomrule
  \end{tabular}
\end{table}

\begin{table}
  \caption{\relic Hyperparameters\vspace{8pt}}
  \label{relic-params}
  \centering
  \begin{tabular}{ll}
    \toprule
    Parameter & Value \\
    \midrule
    Initial reward training steps & 2000 \\
    Reward training steps & 500 \\
    Reward batch size & 500 \\
    Reward learning rate & $3*10^{-4}$ \\
    CL batch size & 50 \\
    CL learning rate & $5*10^{-5}$ \\
    CL embedding dim & 512 \\
    PCA dim & 50 \\
    \tsne perplexity & 30 \\
    \bottomrule
  \end{tabular}
\end{table}

\subsection{Preprocessing and Hyperparameters} 

We train all agents using PPO. The PPO parameters for Reacher are from rl-baselines3-zoo \cite{rl-zoo3}, while the ones for Cheetah and Swimmer are from stable-baselines3 \cite{stable-baselines3}. Agents receive reward feedback every 250000 steps for Reacher and Swimmer, and every 100000 steps for Cheetah. These parameters are kept consistent between \DRLHP and \relic.

To train our contrastive network, we render the Mujoco images and resize them to 100x100. We apply a random crop augmentation to both the anchor and positive example. To form our \tsne representation, we first apply dimensionality reduction with PCA to the concatenation of our contrastive and reward embedding. Then, we run the \tsne algorithm with a perplexity of 30. We show 2000 states to the user each time.

Additional parameters for \DRLHP and \relic are in \Cref{drlhp-params} and \Cref{relic-params} respectively.

\subsection{\DRLHP} 

The initial reward model is trained on states sampled from a random policy, before we begin training our agent. In addition, $25\%$ of the total comparisons are used to train the initial reward model, and comparisons are provided on a hyperbolic schedule in future iterations. 

\subsection{Human timings}\label{appendix:human-timings}

To estimate the amount of time needed to provide a set of cluster rankings, a real human timed themselves for a small number of feedback iterations and used the average. For each iteration, the time measurement starts when the \tsne first loads into the interface and ends when the human submits their cluster orderings.

For \Cref{fig:short-term}, we use the average time per \relic iteration to determine the number of comparisons \DRLHP gets at each iteration. This ensures that each point in \DRLHP has a corresponding point in \relic with the same ``human time'', so we can directly compare the two methods. Instead, for \Cref{fig:long-term} we don't attempt to equalize the amount of human time per-iteration across methods. This enables us to sweep over \DRLHP parameters such as the number of feedback comparisons per iteration that had to be fixed for \Cref{fig:short-term}. By doing this, we can find the best performing hyperparameters without the time-per-iteration constraint. However, this also means that we won't have the same number of datapoints on the x-axis across methods: \DRLHP performs best with many comparisons per iteration, leading to a much higher human-time per iteration relative to \relic.

\subsection{Computational cost}

All experiments were run on an on-premise server. The server has 2 Intel Xeon 6130 32-Core CPUs and 4 NVIDIA GTX 1080 GPUs, shared between multiple projects. The short-term experiments in \Cref{fig:short-term} took around 6 hours of compute. The long-term experiments in \Cref{fig:long-term} took around 60 hours of compute; this is mainly due to running \relic for many iterations to compare against \DRLHP. In practice, one would require far less time (closer to 6 hours) to achieve a reasonable policy.

\section{Additional results}

\begin{figure*}[t]
  \centering
  \includegraphics[width=\textwidth]{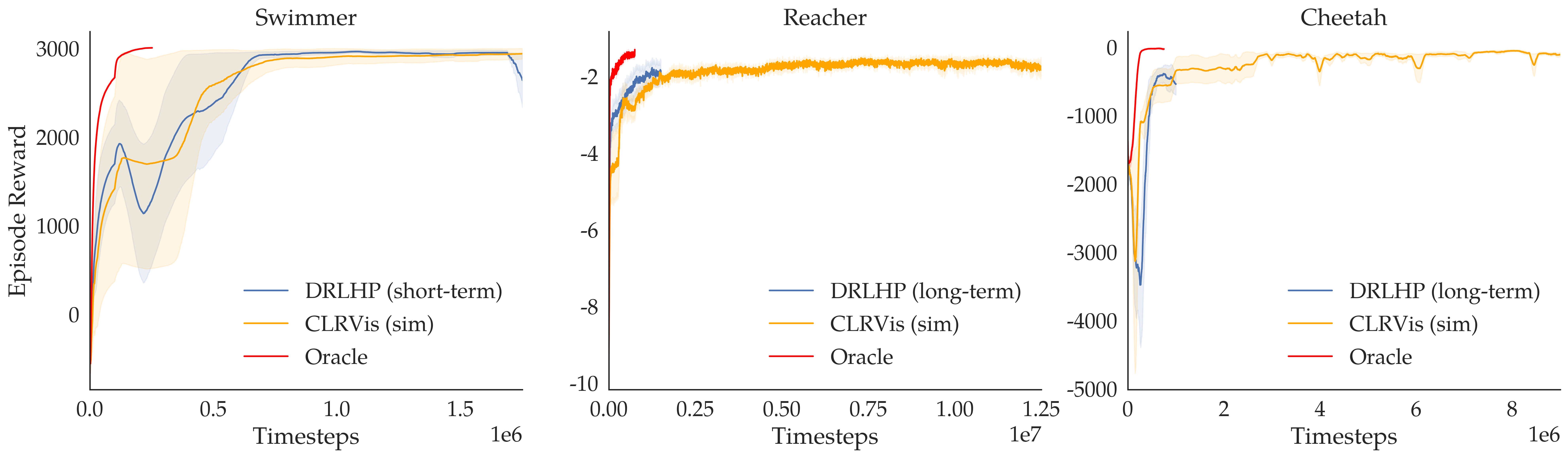}
  \caption{\textbf{Mean episode reward vs timesteps.} Instead of using human time for our x-axis, we switch to environment timesteps. Since \relic uses less human time per timestep, we are able to run it for longer than \DRLHP while spending the same amount of human time.}
  \label{fig:timesteps}
\end{figure*}

\prg{Episode reward in timesteps.} In \Cref{fig:timesteps}, we show how \relic and \DRLHP perform after a certain number of timesteps. Since \relic uses less human time per timestep, we run it for longer than \DRLHP. While \relic might require more timesteps to converge in reward, our goal is to minimize the amount of humantime required instead.

\end{document}